\documentclass[journal]{IEEEtran}
\pdfoutput=1
\usepackage[T1]{fontenc}
\usepackage{amsmath,amssymb,booktabs,graphicx,array,tabularx,microtype,cite}
\usepackage[hidelinks]{hyperref}
\usepackage{xurl}
\graphicspath{{figures/}}
\hypersetup{pdftitle={Scoliosis5K: Multimodal Gait Analysis via Clip-Conditioned RGB-Silhouette Fusion},pdfauthor={Zizhao Peng, Zihan Wang, Zheng Lv, Yan Wang, Ping Li, Fengwei An}}
\newcommand{\R}{\mathbb R}
\newcommand{\unit}{\operatorname{unit}}
\begin{document}
\bstctlcite{tmmrefstyle}
\title{Scoliosis5K: Multimodal Gait Analysis via Clip-Conditioned RGB-Silhouette Fusion}
\author{Zizhao Peng, Zihan Wang, Zheng Lv, Yan Wang, Ping Li, and Fengwei An%
\thanks{Zizhao Peng, Zihan Wang, and Ping Li are with the Department of Computing, The Hong Kong Polytechnic University, Kowloon, Hong Kong (e-mail: 21051291r@connect.polyu.hk; 20031683r@connect.polyu.hk; p.li@polyu.edu.hk).}
\thanks{Zheng Lv is with the Department of Rehabilitation, Longgang District Central Hospital of Shenzhen, Shenzhen, China (e-mail: lzlgzxyy@163.com).}
\thanks{Yan Wang is with the Rehabilitation Medicine Center, The Affiliated Taian City Central Hospital of Qingdao University, Taian, China (e-mail: 13562898274@163.com).}
\thanks{Fengwei An is with the School of Microelectronics, Southern University of Science and Technology, Shenzhen, China (e-mail: anfw@sustech.edu.cn). Corresponding authors: Ping Li and Fengwei An.}}
\maketitle

\begin{abstract}
Color walking video records body boundary motion alongside clothing folds, surface appearance, and recording-environment cues. To examine what an explicit binary silhouette contributes to participant-level classification of scoliosis-related gait when a classifier already observes the paired color sequence, we construct Scoliosis5K, a multi-site collection of 4,998 corridor-walking sequences across 3,010 participant identifiers in which RGB person crops, binary silhouettes, and two-dimensional poses share each recording identifier and frame index. Our architecture uses each patch's pixel deviation from its clip mean to weight two low-rank query and value updates inside a pretrained RGB encoder, fusing the resulting video representation with a silhouette transformer sequence and regional kernel ridge scores extracted from one native-resolution RGB frame per window. Across 988 participants in five participant-disjoint cross-validation folds and three training seeds, the complete system reaches 83.77\% accuracy and 79.91\% macro-averaged F1, compared with 83.27\% accuracy and 79.56\% macro-F1 when the video RGB encoder remains frozen under identical inputs. Zeroing the silhouette vector at fixed weights shifts overall accuracy by only 0.10 percentage points while raising positive-category recall from 68.52\% to 77.78\% and lowering recall for both remaining categories. Macro-averaged AUROC is 0.950 across the cohort and 0.879 among participant pairs matched on acquisition site and recording-count group.
\end{abstract}


\section{Introduction}
\IEEEPARstart{C}{lassifiers} with nearly identical overall accuracy can still identify different subsets of participants when combining binary body outlines with color walking video. Binary silhouettes isolate moving contour geometry, whereas color frames retain torso surface appearance, garment dynamics, and visual cues tied to the recording environment. Determining how each stream shapes a decision requires synchronized observations of the same walk, yet existing public benchmarks such as Scoliosis1K release silhouettes and two-dimensional poses without paired RGB frames~\cite{zhou2024gait,scoliosis1kproject}. We address this requirement with Scoliosis5K, a non-overlapping cohort collected under the same corridor-walking protocol that pairs a native RGB person crop with a binary silhouette and 2D pose landmarks at every indexed frame position.

Applied frame by frame, a frozen image backbone assigns the same patch embedding regardless of how that patch moves across a stride. We condition a pretrained RGB encoder on local clip dynamics by measuring how far the pixels inside each patch deviate from their temporal mean within the window; each patch's deviation relative to the window average sets a zero-parameter gate that mixes two low-rank query and value updates across all attention blocks. In parallel, a divided space--time transformer encodes the paired silhouette sequence and regresses a centered descriptor from the frozen RGB clip during training. Because 24-frame video crops operate at reduced spatial resolution, a second RGB branch compares eight grid regions from one higher-resolution native frame against training participants through a participant-balanced kernel ridge classifier in which every training person contributes equal weight regardless of recording count.

We evaluate 13 silhouette, optical-flow, RGB, and multimodal baselines alongside controlled ablations on five shared participant-disjoint folds ($N=988$, three seeds), comparing our clip-conditioned encoder directly against a frozen RGB video encoder under identical inputs. Two follow-up analyses then trace how predictions behave across the same participants: zeroing the silhouette vector in the fitted classifier shifts overall accuracy by 0.10 percentage points while shifting positive-category recall by 9.26 points, and restricting pairwise evaluation to participants sharing an acquisition site and recording-count group measures how much discriminative ranking persists within matched recording conditions.

\section{Related Work}
\subsection{Gait Representations and Cross-Modal Supervision}
Large-scale recognition corpora such as GREW pair silhouettes with joint trajectories to identify individual walkers across viewpoints~\cite{zhu2021grew}, whereas scoliosis-related gait analysis targets locomotor patterns shared across individuals within a clinical category. Canonical silhouette recognition models aggregate unordered frame sets in GaitSet~\cite{chao2019gaitset}, short-range horizontal strip motion in GaitPart~\cite{fan2020gaitpart}, and combined global-local 3D convolutions in GaitGL~\cite{lin2021gaitgl}. On Scoliosis1K, ScoNet-MT jointly optimizes identity triplet and three-class scoliosis objectives on silhouettes~\cite{zhou2024gait}, DRF augments skeleton maps with posture-asymmetry descriptors~\cite{zhou2025drf}, TG-MILNet pairs temporal bags with clinical text prompts~\cite{li2025tgmil}, and ScoliGaitX combines silhouettes, optical flow, and gait-energy residuals~\cite{vishwakarma2026scoligaitx}. Aguilar-Ortega et al.\ report stronger Scoliosis1K classification when computing dense optical flow on full frames prior to person cropping~\cite{aguilar2026denseflow}, FoundationGait transfers silhouette encoders pretrained across multiple gait datasets~\cite{ye2026foundationgait}, UMGRL models uncertainty across silhouette and skeleton streams~\cite{li2026umgrl}, and Chen et al.\ combine kinematic pose priors with text-guided latent attention pooling~\cite{chen2026clinicalprior}. To benchmark visual gait classification directly on Scoliosis5K, we re-train representative silhouette, optical-flow, RGB, and multimodal architectures on identical participant-disjoint folds.

Multimodal gait architectures also couple outlines with appearance or pose through skeleton-guided feature refinement~\cite{wang2024gaitref} or color-derived semantic body parts as in GaitParsing~\cite{wang2024gaitparsing}. Rather than relying on part segmentation or anatomical landmarks, our native-frame branch compares appearance descriptors on a fixed $4\times 2$ grid and tests spatial correspondence directly against an all-pairs region kernel. For sequence fusion, directional cross-modal attention (MulT) queries one feature stream from another~\cite{tsai2019multimodal}, whereas cross-modal distillation transfers supervision across paired views during training~\cite{gupta2016crossmodal}. In our model, both RGB and silhouette streams feed the classifier at inference while the silhouette branch regresses a centered target from the synchronized RGB clip during training, allowing donor-swap controls that hold classifier inputs fixed while varying participant correspondence in the teacher target.

\subsection{Adapting Image Backbones to Gait Video}
Pretrained vision backbones such as DINOv2~\cite{oquab2024dinov2} and Perception Encoder~\cite{bolya2025pe} provide dense patch representations that BigGait adapts for appearance-based gait recognition~\cite{ye2024biggait}, complementing self-supervised video backbones such as VideoMAE~\cite{tong2022videomae}. Transferring image encoders to video typically introduces static low-rank weight increments (LoRA)~\cite{hu2022lora}, depthwise spatiotemporal adapters (ST-Adapter)~\cite{pan2022st}, or learned router mixtures over low-rank experts (X-LoRA)~\cite{buehler2024xlora}. Instead of learning a parameterized router, our adapter weights two low-rank updates at every block (including the first) using each patch's observed pixel $L_1$ deviation relative to the rest of the clip. Permuting the frames permutes their gate values identically; because the RGB attention pool omits temporal position encodings, its pooled vector remains permutation-invariant while the ordered silhouette transformer retains frame order.

\section{The Scoliosis5K Dataset}
\label{sec:dataset}
\subsection{Walking Recordings and Synchronized Modalities}
\begin{figure*}[!t]
\centering
\includegraphics[width=\textwidth]{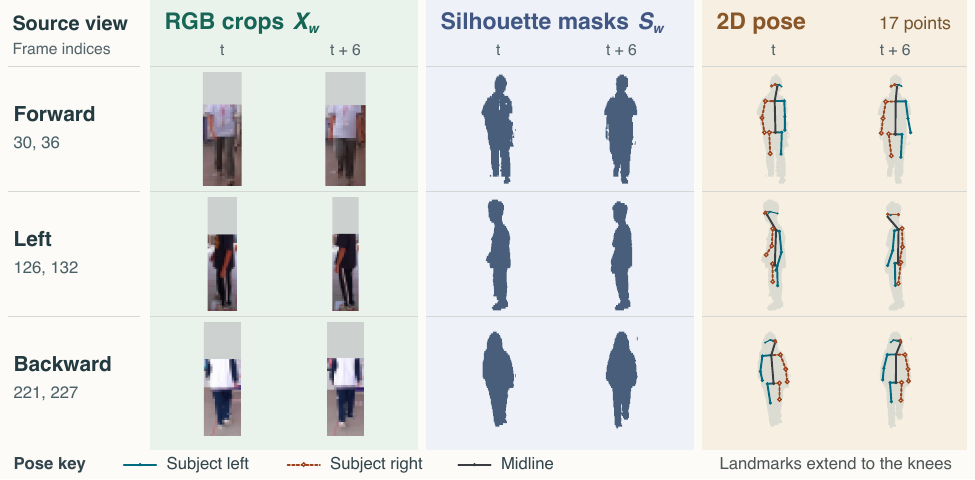}
\caption{Synchronized modalities from two frames ($t$ and $t+6$) of three training participants across forward, left, and backward walking views. Zero-based frame indices appear at left. RGB crops ($X_w$) are block-averaged and masked through the shoulders for display; at right, 17 stored ViTPose landmarks (indices 0--16 of the 25-keypoint COCO-25 layout, extending through the knees; solid lines with filled markers denote subject left, dashed lines with open markers denote subject right) overlay the paired $64\times 64$ silhouette ($S_w$).}
\label{fig:dataset_modalities}
\end{figure*}

Scoliosis5K follows the corridor-walking protocol of Scoliosis1K~\cite{zhou2024gait} while excluding all acquisitions and participant identifiers matched to Scoliosis1K (which comprised 1,493 recordings from 1,050 participants without RGB imagery). Collected across nine sites, Scoliosis5K contains 4,998 sequences linked to 3,010 participant identifiers, with 300 indexed positions per sequence (1,499,400 frame positions in total). Every indexed position links a native RGB person crop, a $64\times64$ binary silhouette, and 17 two-dimensional pose landmarks extracted by ViTPose (indices 0--16 of the 25-keypoint COCO-25 layout, spanning the head, torso, arms, hips, and knees) with confidence scores via a shared recording identifier and frame index (Fig.~\ref{fig:dataset_modalities}). Displayed RGB examples in Figs.~\ref{fig:dataset_modalities} and~\ref{fig:architecture} apply head masking through the shoulders and block averaging for manuscript display privacy, whereas all models in this study operated on the de-identified corridor crops.

\begin{table*}[!t]
\caption{Recording Counts and Participant Characteristics of the Full Scoliosis5K Collection. Bracketed Counts Indicate Participants With Recorded Measurements.}
\label{tab:dataset}\centering\small\setlength{\tabcolsep}{3pt}
\begin{tabular*}{\textwidth}{@{\extracolsep{\fill}}lrrrr@{}}
\toprule
Characteristic & Overall & Negative & Neutral & Positive\\
\midrule
Recordings by screening label & 4,998 & 1,666 & 1,666 & 1,666\\
\midrule
Participants & 3,010 & 1,639 & 882 & 485\\
Sex (female / male) & 1,644 / 1,366 & 854 / 785 & 465 / 417 & 322 / 163\\
Age (years) & 14.82 $\pm$ 2.40 [3,000] & 15.02 $\pm$ 1.94 [1,629] & 14.51 $\pm$ 1.99 [882] & 14.72 $\pm$ 3.95 [485]\\
Height (cm) & 160.56 $\pm$ 10.64 [3,008] & 161.32 $\pm$ 10.71 [1,639] & 160.25 $\pm$ 10.77 [880] & 158.58 $\pm$ 9.89 [485]\\
Weight (kg) & 48.73 $\pm$ 12.10 [3,010] & 50.48 $\pm$ 12.50 [1,639] & 47.58 $\pm$ 11.94 [882] & 44.90 $\pm$ 9.67 [485]\\
Body mass index (kg/m$^2$) & 18.65 $\pm$ 3.13 [3,008] & 19.16 $\pm$ 3.36 [1,639] & 18.21 $\pm$ 2.73 [880] & 17.69 $\pm$ 2.60 [485]\\
\bottomrule
\end{tabular*}
\par\smallskip
\begin{minipage}{\textwidth}\footnotesize Recordings retain their clinical screening labels, and each participant is counted once overall. Category-specific participant columns omit four individuals whose 13 recordings span multiple labels. Continuous entries report mean $\pm$ sample standard deviation.\end{minipage}
\end{table*}

\subsection{Participants, Labels, and Five-Fold Evaluation Cohort}
Scoliosis5K follows the acquisition, clinical screening, and data-recording standards established by Scoliosis1K~\cite{zhou2024gait}, assigning each walking sequence to one of three screening categories (negative, neutral, or positive), with full-corpus demographics summarized in Table~\ref{tab:dataset} using the first recorded value per field in index order. To support controlled multi-seed benchmarking across diverse video and gait architectures on a recording-balanced cohort without cross-label ambiguity, our participant-disjoint five-fold evaluation cohort comprises 988 single-label participants contributing 1,628 recordings across all nine sites (548 negative, 550 neutral, and 530 positive walks; 531 female and 457 male; median age 15 years, interquartile range 13--16 among the 986 with recorded age; Supplementary Table~IV). Within this cohort, 569 participants contributed one walk, 303 contributed two, and 116 contributed three or more; site sizes range from 3 to 295 participants, and four sites lack at least one category. Because individual radiographic Cobb angles are not attached to the video records, the three screening labels are evaluated as nominal categories: 530 positive recordings come from 162 participants (3.27 walks per person on average), 550 neutral recordings from 289 participants, and 548 negative recordings from 537 participants (1.02 walks per person). Both fold assignment and evaluation metrics operate strictly at the participant level.

\section{Method}
\subsection{Dual Visual Streams Across Spatial Scales}
\begin{figure*}[!t]
\centering
\includegraphics[width=\textwidth]{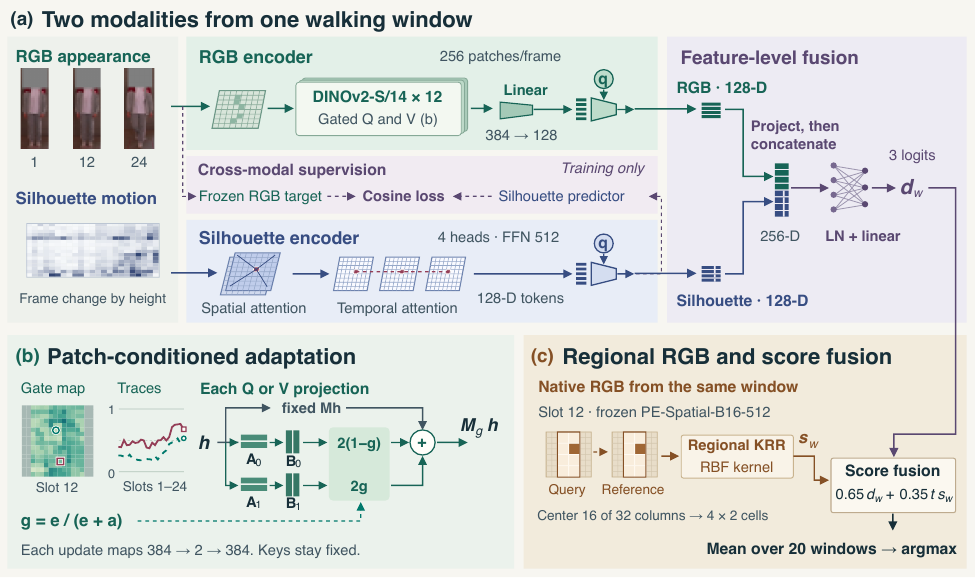}
\caption{Overview of paired RGB and silhouette gait classification.
(a) Slots 1, 12, and 24 from one training walk sampled every second frame; RGB thumbnails retain native crop aspect ratios with head masking and block averaging for display, while the silhouette strip map plots mean absolute frame-to-frame change across 16 height bands (the encoder receives the binary masks directly). Stream-specific learned queries pool valid tokens, a training-only predictor aligns the pooled silhouette vector with the centered frozen RGB target, and the classifier concatenates both projected streams to output window logits $d_w$.
(b) Patch-deviation gate map and temporal traces computed on the model RGB inputs prior to display masking across the 24 slots; each patch gate $g_{tr}$ mixes two rank-two updates in every query and value projection while key projections remain fixed.
(c) Native-frame regional kernel ridge regression (KRR) compares eight central grid cells from slot 12 against training participants, producing scores $s_w$ scaled by $t$ and fused with $d_w$ across 20 windows per participant.}
\label{fig:architecture}
\end{figure*}

Figure~\ref{fig:architecture} traces a 24-frame evaluation window $w$ through our two-branch classifier. Each window aligns RGB crops $X_w$ and binary silhouettes $S_w$ over a shared recording and frame-index interval, with 24 frame slots and stream-specific validity masks. The paired video network outputs three class logits $d_w$, while a complementary estimator scores one native-resolution RGB crop from the same window on a fixed spatial grid to produce three regional scores $s_w$; neither path requires anatomical landmark detection.

\subsection{Conditioning Patch Projections on Clip Appearance Deviation}
Along the RGB video path, person crops are resized to $96\times64$, resampled to $224\times149$, and padded horizontally to $224\times224$ before entering DINOv2 ViT-S/14~\cite{oquab2024dinov2} on a $16\times16$ patch grid. Prior to channel normalization, let $x_{tp}\in[0,1]^3$ denote the RGB pixel at valid spatial location $p$ in valid frame $t$, with temporal mean $\bar x_p$ across valid frames in the window. Over the non-padded pixels $P_r$ inside patch $r$, the patch deviation is
\begin{equation}
 e_{tr}=\frac{1}{3|P_r|}\sum_{p\in P_r}\lVert x_{tp}-\bar x_p\rVert_1.
 \label{eq:dev}
\end{equation}
Denoting by $a$ the mean of $e_{tr}$ across all valid frames and nonempty patches in the window (excluding padded frames from both means), we define the patch gate as $g_{tr}=e_{tr}/(e_{tr}+a)$ when $a>0$, and set $g_{tr}=1/2$ for constant clips, fully padded patches, and the class token.

Inside each of the 12 transformer blocks, let $M$ denote a frozen pretrained query or value weight matrix. At a token with hidden state $h$ and gate $g$, its adapted linear projection is
\begin{align}
 M_g h &= Mh+2\big[(1-g)U_0 h+g U_1 h\big],\label{eq:adapter}\\
 U_j &= B_j A_j,\quad
 A_j\in\R^{2\times384},\quad B_j\in\R^{384\times2},\nonumber
\end{align}
where pretrained biases, backbone weights, and key projections stay frozen, each query or value matrix owns dedicated factor pairs $(A_j,B_j)$ initialized with $B_j=0$ (73,728 trainable adapter weights across all blocks), and a static-gate control fixes $g=1/2$ everywhere (spanning the same subspace as a single rank-four update). On any nonempty patch in a varying window, the update in~\eqref{eq:adapter} decomposes into a static component and a zero-crossing contrast term:
\begin{equation}
 (U_0+U_1)h + \frac{e_{tr}-a}{e_{tr}+a}(U_1-U_0)h .
 \label{eq:contrast}
\end{equation}
Because the second coefficient changes sign as $e_{tr}$ crosses the window mean $a$, identical single-frame patch features receive distinct projections depending on clip context. Setting $v=e_{tr}/a$ with $\langle v\rangle=1$ over valid patch--frame pairs gives the expected gate identity
\begin{equation}
 \langle g\rangle=\frac{1}{2}-\frac{1}{4}\left\langle\frac{(v-1)^2}{1+v}\right\rangle ,\label{eq:gate_mean}
\end{equation}
which equals $1/2$ only under uniform deviation and drops below $1/2$ as motion concentrates spatially. Uniform positive pixel rescaling leaves $g_{tr}$ invariant because $e_{tr}$ and $a$ scale together, whereas within-clip lighting shifts or camera motion still modulate $e_{tr}$.

\subsection{Joint Classification With Paired Silhouette Distillation}
Per valid frame, the adapted DINOv2 backbone yields 256 patch tokens of width 384, which a learned linear layer maps to 128 dimensions. In parallel, each $64\times64$ silhouette frame is partitioned into 64 non-overlapping $8\times8$ patches, projected to width 128, and encoded by a divided space--time transformer block applying spatial self-attention within frames followed by temporal self-attention per patch location (four heads, pre-normalization, 512-wide feed-forward network, and fixed sinusoidal spatial and temporal encodings). Separate learned queries pool valid tokens from each stream; the two pooled vectors are projected, concatenated, layer-normalized, and mapped by an affine head to window logits $d_w$.

During training, an unadapted DINOv2 pass also computes a frozen clip target $r_w$ by averaging all 256 RGB patch tokens across valid frames in the window. With $\mu$ denoting the mean of $r_w$ over candidate windows in the current training fold and $q(p_w^S)\in\R^{384}$ denoting an auxiliary projection of the pooled silhouette vector $p_w^S$, we minimize
\begin{equation}
 \mathcal L_w=\mathrm{CE}(d_w,y_i)+1-
 \unit\!\big(q(p_w^S)\big)^{\!\top}\unit(r_w-\mu),
 \label{eq:loss}
\end{equation}
where $\unit(v)=v/\lVert v\rVert_2$. The teacher target comes from the same participant and window, remains fixed throughout optimization, and is discarded at inference.

\subsection{Participant-Balanced Native-Frame Regional Kernel Classifier}
For the complementary high-resolution RGB path, we select one-based slot 12 of the window (or the nearest valid slot, breaking ties toward the earlier frame), resize the native crop to $512\times512$, and extract a $32\times32\times768$ patch grid from frozen PE-Spatial-B16-512~\cite{bolya2025pe}. Retaining the central 16 columns, we partition them into four horizontal rows by two vertical columns, average the $8\times8$ tokens per cell, and $L_2$-normalize each cell to obtain eight regional descriptors $\{u_{wr}\}_{r=1}^8$. Across windows $w$ and $v$, regional similarity is measured by
\begin{equation}
 k(w,v)=\frac{1}{8}\sum_{r=1}^8
 \exp\!\left[-\lVert u_{wr}-u_{vr}\rVert_2^2/b\right],
 \label{eq:kernel}
\end{equation}
where bandwidth $b\ge 10^{-4}$ is the median squared Euclidean distance between corresponding cells across distinct training participants, estimated on the first valid window of up to 128 randomly sampled training participants using a fixed random seed. An all-pairs control kernel averages over all 64 cell pairs without spatial correspondence; writing $\phi_{wr}$ for the radial basis feature of cell $r$ and $\bar\phi_w=\tfrac18\sum_r\phi_{wr}$, the all-pairs kernel equals $\langle\bar\phi_w,\bar\phi_v\rangle$, so subtracting it from~\eqref{eq:kernel} isolates cell-specific deviations from each image mean:
\begin{equation}
 \frac{1}{8}\sum_{r=1}^{8}\langle\phi_{wr}-\bar\phi_w,\phi_{vr}-\bar\phi_v\rangle.\label{eq:regiondiff}
\end{equation}

To prevent multi-walk participants from dominating the reference set, kernel ridge regression fits one target per training participant. Let $V_j$ collect up to two valid training windows (the first two distinct recording and start-index pairs in candidate order) for participant $j$. Defining the participant kernel query and training Gram matrix by
\begin{align}
 k_j(w)&=|V_j|^{-1}\sum_{v\in V_j}k(w,v),\nonumber\\
 K_{ij}&=|V_i|^{-1}\sum_{w\in V_i} k_j(w),\label{eq:gram}
\end{align}
we center both $K$ and the one-hot label matrix $Y$ across training participants to form $(K_c,Y_c)$ and solve $(K_c+\lambda I)^{-1}Y_c$. Query vectors are centered with the training statistics before adding the training class prior; because this readout is affine in the query kernel vector, averaging window scores $s_w$ over a participant's 20 windows equals evaluating their mean kernel vector once.

\subsection{Window Score Fusion and Participant Prediction}
Scaling regional scores by the fold-specific inverse temperature $t$, we fuse both branches per window and average across a participant's 20 evaluation windows:
\begin{equation}
 z_w=0.65\,d_w+0.35\,t\,s_w,\quad
 \hat y_i=\arg\max_c \frac{1}{20}\sum_{w=1}^{20} z_{iw,c}.
 \label{eq:fusion}
\end{equation}
If a window lacks a valid regional frame, $z_w=d_w$ (all 19,760 held-out slots had valid regional frames, with repeated slots retained at sampled multiplicity). Each training seed yields an independent set of participant predictions.

\section{Experimental Design}
\subsection{Evaluation Unit and Anti-Leakage Protocol}
All sequences from a participant carry the same nominal screening label and reside in the same participant-disjoint fold, yielding five held-out folds of 199, 199, 197, 197, and 196 participants. Each neural model is trained across three random seeds (73111, 73112, and 73113); within each seed, held-out predictions from all five folds are pooled across the 988 participants to compute accuracy and macro-averaged F1, and tables report the mean $\pm$ population standard deviation across the three seeds (so each participant contributes up to three decisions in pooled counts). Architectural hyperparameters were developed on a 631-train / 158-validation split within the first outer training fold, and the $0.65/0.35$ fusion weights in~\eqref{eq:fusion} were locked on the frozen-video plus regional model and retained unchanged when substituting the adapted video branch.

\subsection{Training Protocol, Controls, and Baseline Adaptations}
Window banks are built by sampling a participant's recording uniformly and then a valid 24-frame start index uniformly with replacement, locking eight training candidates and 20 evaluation windows per participant across paired video configurations and seeds. Paired video models train for 60 epochs at batch size eight (one candidate window per participant per epoch; 5,940 updates per outer fold) with AdamW~\cite{loshchilov2019decoupled}, 5\% warm-up, cosine decay, gradient clipping at 1.0, dropout 0.1 on the silhouette transformer and attention pools, matrix weight decay 0.01 (zero on biases and normalizations), learning rate $3\times10^{-4}$ for the silhouette path and fusion head, and $10^{-4}$ for the image adapter, reloading final weights for evaluation (15 neural fits plus five fold-level regional classifiers). Within each outer training fold, stratified five-fold cross-validation selects $\lambda\in\{0.01,0.1,1,10\}$ (estimating bandwidth and centering inside each inner split and fitting $t\in[0.25,16]$ on participant-averaged inner logits via multiclass cross-entropy, breaking ties toward larger $\lambda$) before refitting on the full outer fold; every outer fold selected $\lambda=0.1$ with $t\in[3.87,4.23]$ (Supplementary Table~V).

Controlled comparisons isolate each component: freezing the RGB video encoder with and without regional KRR; comparing patch-varying vs.\ constant ($g=1/2$) gates at matched initialization, batches, and a 30-epoch schedule on the 631/158 development split (seeds 73112 and 73113); and fitting the 64-pair position-agnostic kernel under identical inner-fold selection. We also re-train published gait baselines on the identical 988-participant folds and 20 evaluation windows: silhouette-based GaitSet, GaitPart, GaitGL, ScoNet-MT, and FoundationGait~\cite{ye2026foundationgait} (both the probe-then-fine-tune 0.03B model, the strongest Scoliosis1K variant in Ye et al., and linear-probed and fine-tuned 0.13B models); silhouette-derived motion adaptations of TG-MILNet and ScoliGaitX; ScoNet-MT on two-channel cropped RGB RAFT flow (contrasting with full-frame three-color OF-ScoNet-MT~\cite{aguilar2026denseflow}); and RGB/multimodal DINOv2 linear probing, VideoMAE, and MulT (Supplementary Section~VIII details all baseline training schedules and input preprocessing). Paired 95\% accuracy intervals draw 10,000 participant-level bootstrap resamples coupling all three seed decisions per participant (Supplementary Tables~VI--VIII report per-seed counts, class recalls, and confusion matrices).

\section{Results}
\label{sec:results}
\subsection{Participant-Level Classification}
\begin{table*}[!t]
\caption{Three-Category Participant-Level Gait Classification on Scoliosis5K ($N=988$ Participants Pooled Across Five Participant-Disjoint Folds, Mean $\pm$ Population SD Across Three Seeds, \%). Baselines Are Re-Trained on Identical Scoliosis5K Folds.}
\label{tab:main}\centering\small
\begin{tabular*}{\textwidth}{@{\extracolsep{\fill}}llrr@{}}
\toprule
System & Classifier Inputs & Accuracy (\%) & Macro-F1 (\%)\\
\midrule
\textbf{Clip-conditioned video + regions} & RGB, silhouette, regions & \textbf{83.77} $\pm$ 0.41 & \textbf{79.91} $\pm$ 0.56\\
Clip-conditioned video & RGB, silhouette & 83.64 $\pm$ 0.24 & 79.82 $\pm$ 0.32\\
Frozen-encoder video + regions & RGB, silhouette, regions & 83.27 $\pm$ 0.13 & 79.56 $\pm$ 0.07\\
Frozen-encoder video & RGB, silhouette & 83.00 $\pm$ 0.30 & 79.30 $\pm$ 0.28\\
\midrule
GaitSet~\cite{chao2019gaitset} & Silhouette & 67.00 $\pm$ 0.44 & 55.83 $\pm$ 1.20\\
GaitPart~\cite{fan2020gaitpart} & Silhouette & 56.01 $\pm$ 0.78 & 28.15 $\pm$ 2.15\\
GaitGL~\cite{lin2021gaitgl} & Silhouette & 54.59 $\pm$ 0.33 & 24.04 $\pm$ 0.79\\
ScoNet-MT~\cite{zhou2024gait} & Silhouette & 73.65 $\pm$ 0.88 & 69.75 $\pm$ 1.08\\
FoundationGait-0.13B linear probe~\cite{ye2026foundationgait} & Silhouette & 68.45 $\pm$ 0.13 & 63.64 $\pm$ 0.07\\
FoundationGait-0.13B fine-tuned~\cite{ye2026foundationgait} & Silhouette & 75.07 $\pm$ 0.62 & 69.45 $\pm$ 0.83\\
FoundationGait-0.03B (probe then fine-tune)~\cite{ye2026foundationgait} & Silhouette & 77.80 $\pm$ 0.62 & 73.83 $\pm$ 0.65\\
TG-MILNet~\cite{li2025tgmil} & Silhouette, derived flow, text prompts & 56.11 $\pm$ 5.71 & 52.70 $\pm$ 3.63\\
ScoliGaitX~\cite{vishwakarma2026scoligaitx} & Silhouette, derived flow, GEI residuals & 67.95 $\pm$ 2.98 & 64.54 $\pm$ 1.82\\
ScoNet-MT with cropped RGB flow~\cite{zhou2024gait,aguilar2026denseflow} & Two-channel RAFT displacements & 75.07 $\pm$ 1.13 & 70.62 $\pm$ 1.53\\
\midrule
DINOv2 linear probe~\cite{oquab2024dinov2} & RGB & 80.67 $\pm$ 0.17 & 75.79 $\pm$ 0.17\\
VideoMAE~\cite{tong2022videomae} & RGB & 81.82 $\pm$ 0.13 & 77.39 $\pm$ 0.11\\
Directional cross-attention (MulT)~\cite{tsai2019multimodal} & RGB, silhouette & 82.49 $\pm$ 0.08 & 78.55 $\pm$ 0.17\\
\bottomrule
\end{tabular*}
\end{table*}

Across all 988 held-out participants, the complete model reaches 83.77\% accuracy and 79.91\% macro-F1 (Table~\ref{tab:main}; predicting the majority category for every participant yields 54.35\% accuracy and 23.48\% macro-F1). Holding the silhouette and regional inputs fixed while freezing the video RGB encoder yields 83.27\% accuracy and 79.56\% macro-F1, and directional cross-attention (MulT) over the two video streams reaches 82.49\% and 78.55\%. Among silhouette-only methods, FoundationGait-0.03B is the strongest baseline at 77.80\% accuracy and 73.83\% macro-F1; fine-tuning FoundationGait-0.13B improves accuracy from 68.45\% (linear probe) to 75.07\%, while ScoNet-MT reaches 73.65\% on silhouettes and 75.07\% on cropped RGB optical flow.

\subsection{Clip-Conditioned RGB Encoder Adaptation}
Holding the silhouette transformer, distillation objective, and window bank fixed, introducing clip-conditioned low-rank projections into the frozen video reference raises accuracy from 83.00\% to 83.64\% (+19 net correct participant--seed decisions across the three seeds) and macro-F1 from 79.30\% to 79.82\%. On the 631/158 development control at the identical 73,728-weight budget, varying the pixel-deviation gate across patches yields 264 correct decisions out of 316 participant--seed observations versus 263 for a constant gate ($g=1/2$), showing that most of the improvement over the frozen encoder stems from low-rank query--value adaptation, with dynamic patch-level gating adding a modest margin at matched parameter count.

\subsection{Native-Frame Regional Contributions}
Evaluated on its own from a single color frame per window, the participant-balanced regional classifier reaches 82.79\% accuracy and 78.67\% macro-F1. Fusing its scores with the adapted video branch corrects 24 errors, introduces 20 new errors, and shifts seven predictions between incorrect categories, yielding a $+0.13$-point accuracy gain (paired 95\% interval $[-0.34,0.61]$) and $+0.09$-point macro-F1 gain (+12 correct negative decisions, $-8$ neutral decisions, and unchanged positive count across seeds; $+0.78$ accuracy points, $[-0.24,1.86]$, over the frozen video reference without regions). Comparing the corresponding-cell kernel~\eqref{eq:kernel} against the 64-pair position-agnostic kernel on top of the frozen video classifier gives 83.27\% vs.\ 83.30\% accuracy (79.56\% vs.\ 79.50\% macro-F1), indicating that the regional contribution stems primarily from high-resolution appearance similarity rather than strict grid-cell alignment.

\subsection{Cross-Modal Supervision and Fixed-Weight Stream Interventions}
On the frozen video reference, training the silhouette encoder without an RGB teacher gives 78.85\% macro-F1 (82.69\% accuracy), whereas supervising it with the synchronized RGB clip from the same walk raises macro-F1 to 79.30\%. Substituting another participant's RGB clip from the same screening category yields 78.82\%, and regressing the leave-one-out class mean reaches 79.15\%; the instance-paired target exceeds the same-class donor in two of the three seeds, while the class-mean gain shows that shared category structure also guides the silhouette branch.

To examine how the jointly trained classifier relies on each video stream, we replayed all 15 fitted video models across their held-out folds while zeroing either the projected RGB vector or the projected silhouette vector immediately before the shared layer normalization, with and without the single-frame regional scores (all 19,760 windows contained valid RGB frames, silhouettes, and regional descriptors; keeping both streams active reproduced every saved prediction identically).

\begin{table}[!t]
\caption{Fixed-Weight Stream Removal Interventions Across 2,964 Participant--Seed Decisions. Accuracy and Macro-F1 (\%) Are Averaged Across Three Seeds.}
\label{tab:streams}\centering\small\setlength{\tabcolsep}{3pt}
\begin{tabular*}{\columnwidth}{@{\extracolsep{\fill}}llrrr@{}}
\toprule
Video Inputs & Regions & Correct & Acc. (\%) & Macro-F1 (\%)\\
\midrule
RGB + silhouette & Yes & 2,483 & 83.77 & 79.91\\
RGB + silhouette & No & 2,479 & 83.64 & 79.82\\
RGB & Yes & 2,486 & 83.87 & 80.57\\
RGB & No & 2,474 & 83.47 & 80.13\\
Silhouette & Yes & 1,363 & 45.99 & 41.44\\
Silhouette & No & 802 & 27.06 & 22.19\\
\bottomrule
\end{tabular*}
\end{table}

\begin{figure*}[!t]
\centering
\includegraphics[width=\textwidth]{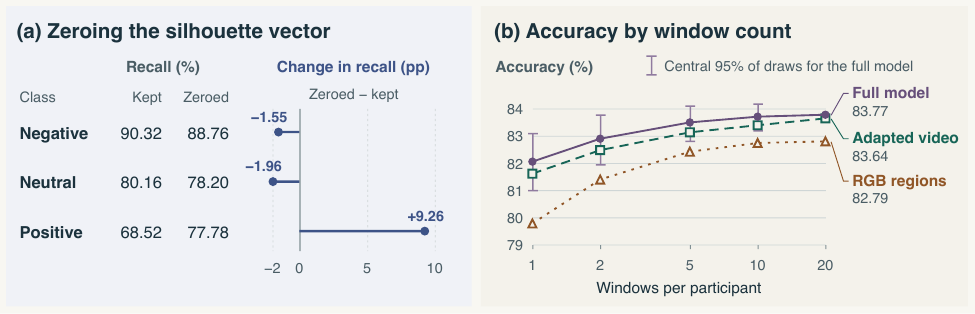}
\caption{Silhouette-removal intervention and multi-window scaling on the 988 held-out participants.
(a) Class-specific recall with the silhouette vector retained or zeroed prior to shared layer normalization (regional RGB scores active in both conditions; recall differences are computed before rounding, in percentage points).
(b) Mean participant accuracy across three seeds and 500 random permutations of $k\in\{1,2,5,10,20\}$ evaluation windows per participant (all models share identical sampled windows per draw; vertical bars span the 2.5th to 97.5th percentiles of the full model across draws, with $k=20$ using the full bank).}
\label{fig:readouts}
\end{figure*}

With regional scores retained, zeroing the silhouette vector alters 123 of 2,964 participant--seed predictions (Table~\ref{tab:streams}), correcting 53 errors, introducing 50 new errors, and shifting 20 predictions between incorrect classes. Overall accuracy moves by $+0.10$ percentage points (paired 95\% interval $[-0.78,0.98]$) and macro-F1 moves from 79.91\% to 80.57\%, yet the error profile shifts markedly across categories: positive recall climbs from 68.52\% to 77.78\% ($+9.26$ points), whereas negative recall declines from 90.32\% to 88.76\% and neutral recall from 80.16\% to 78.20\% (Fig.~\ref{fig:readouts}a and Supplementary Table~XI). Conversely, zeroing the RGB video vector reduces accuracy to 45.99\% even with regional RGB scores retained (27.06\% without regions), reflecting the scale shift at the shared layer normalization after joint training.

\subsection{Number of Evaluation Windows per Participant}
To measure sensitivity to the number of observed clips per participant, we drew 500 random permutations of each participant's 20 evaluation slots and scored the first $k\in\{1,2,5,10,20\}$ entries, sharing identical sampled slots across all models and seeds within each draw (because slots are sampled across a person's recordings, $k$ indexes distinct 24-frame windows rather than a contiguous walk duration). From a single window ($k=1$), the full model averages 82.05\% accuracy (central 95\% draw range 81.00--83.10\%; Fig.~\ref{fig:readouts}b), rising to 83.49\% at five windows, 83.70\% at ten windows (central 95\% range 83.20--84.18\%), and 83.77\% across all twenty windows. Over the same $k=1$ to $k=20$ range, the single-frame regional classifier gains 3.02 accuracy points, classifying 818 of 988 participants correctly with the full 20-window bank.

\subsection{Ranking Within Matched Acquisition Conditions}
Because the nine collection sites differ in both category prevalence and walks recorded per participant, we restricted one-versus-rest AUROC comparisons to participant pairs sharing both acquisition site and recording-count group (one, two, or three-plus walks). Out of 578,010 class-specific pairs, 15,126 (2.62\%) satisfied both criteria; macro-AUROC was 0.950 across the full cohort and 0.879 within matched groups, with the positive category having the fewest matched pairs (2,365) and a matched AUROC of 0.819 versus 0.943 overall (Supplementary Table~IX).

\section{Discussion}
Both RGB branches agree on the majority of participants: fusing high-resolution single-frame regional scores with the adapted video stream alters 51 of 2,964 participant--seed decisions (24 corrections and 20 new errors). Near-parity between the corresponding-cell and all-pairs regional kernels likewise cautions against interpreting fixed grid cells as anatomical landmarks, since body parts shift across grid cells with walking phase and viewpoint, and illumination or camera motion can elevate patch temporal deviations alongside limb movement.

Retaining the explicit silhouette stream reshapes which participants the jointly fitted classifier detects rather than shifting headline accuracy. Removing the silhouette vector adds 45 net correct positive participant--seed decisions while losing 42 correct decisions across the negative and neutral categories; removing the RGB video vector before shared layer normalization alters feature scale and lowers joint accuracy to 45.99\%, whereas the standalone regional RGB classifier reaches 82.79\%. Multi-site clinical collection also introduces site-level variation in category prevalence and repeat-walk frequency, and background cues remain partially visible in corridor crops. Restricting pairwise evaluation to participants matched on both acquisition site and recording-count group confirms that substantial discriminative ranking persists under matched conditions (macro-AUROC 0.879 vs.\ 0.950 overall). Evaluating generalization to unseen clinical centers, correlating gait scores with radiographic Cobb angles, and benchmarking models trained directly under privacy-preserving head masking remain important directions for future study.

\section{Conclusion}
We introduced Scoliosis5K to align RGB crops, binary silhouettes, and 2D pose landmarks at shared frame indices across 4,998 walking sequences from 3,010 participants. Combining clip-conditioned low-rank RGB adaptation, paired RGB-to-silhouette distillation, and participant-balanced native-frame regional kernel fusion reached 83.77\% accuracy, 79.91\% macro-F1, and 0.950 macro-AUROC (0.879 within site- and count-matched pairs) across 988 held-out participants. Controlled stream interventions show that while paired RGB features drive overall accuracy, the silhouette branch substantially shifts sensitivity between positive and non-positive screening categories, moving positive recall by more than nine percentage points when overall accuracy changes by one tenth of a point.

\section*{Ethics Statement}
The acquisition protocol and secondary analysis of the Scoliosis5K recordings were approved by the institutional ethics committees of the participating clinical centers, with informed consent obtained from all participants and their legal guardians. The present study analyzed existing de-identified recordings without new participant recruitment, contact, or clinical intervention.

\section*{Acknowledgment}
The authors thank the Department of Rehabilitation, Longgang District Central Hospital of Shenzhen, and the Rehabilitation Medicine Center, The Affiliated Taian City Central Hospital of Qingdao University, for clinical support and data collection. AI tools were used to assist with English language polishing, research and ideation, identifying related work, and accelerating code development and visualization; the authors assume full responsibility for all scientific content, experimental protocols, and results.

\bibliographystyle{IEEEtran}
\bibliography{references}

\clearpage
\twocolumn[%
\begin{center}
{\LARGE Supplementary Material for\\[0.25em] \emph{Scoliosis5K: Multimodal Gait Analysis via Clip-Conditioned RGB-Silhouette Fusion}\par}
\vspace{0.8em}
{\large Zizhao Peng, Zihan Wang, Zheng Lv, Yan Wang, Ping Li, and Fengwei An\par}
\vspace{1.4em}
\end{center}
]
\setcounter{section}{0}
\renewcommand{\theHsection}{supp.\arabic{section}}
\setcounter{table}{3}
\setcounter{figure}{3}

\section{Paired Observations in Scoliosis5K}
Each recording in Scoliosis5K contains 300 WebP person crops numbered 00000 to 00299, linked to the paired $64\times 64$ binary silhouettes and 17 stored 2D pose landmarks extracted by ViTPose (indices 0--16 of the 25-keypoint COCO-25 layout, spanning the head, torso, arms, hips, and knees) via a shared pseudonymous sequence identifier, frame index, and clinical screening label. A fixed participant-disjoint roster maps the 1,628 study recordings from the 988 single-label participants to five evaluation folds.

\section{Fold Composition and Regional Fits}
Whereas the full Scoliosis5K collection includes four individuals whose 13 recordings span multiple screening labels, each of the 988 benchmark participants carries a single consistent label across all linked recordings. Table~\ref{tab:cohortdemo} summarizes participant demographics across the 988-person cohort using the first recorded value per field in index order; height, weight, and BMI varied slightly across repeat visits for six individuals.

\begin{table}[!t]
\caption{Demographics of the Five-Fold Evaluation Cohort by Screening Category (Participant Mean $\pm$ Sample SD). Two Negative Participants Lacked a Recorded Age; All Other Fields Were Complete.}
\label{tab:cohortdemo}\centering\footnotesize\setlength{\tabcolsep}{1pt}
\begin{tabular*}{\columnwidth}{@{\extracolsep{\fill}}lrrrr@{}}
\toprule
 & Overall & Negative & Neutral & Positive\\
\midrule
Recordings & 1,628 & 548 & 550 & 530\\
Participants & 988 & 537 & 289 & 162\\
Female / male & 531/457 & 267/270 & 152/137 & 112/50\\
Age (years) & $14.8\!\pm\!2.4$ & $15.0\!\pm\!2.0$ & $14.6\!\pm\!2.0$ & $14.5\!\pm\!3.9$\\
Height (cm) & $160.8\!\pm\!10.7$ & $161.8\!\pm\!10.4$ & $160.8\!\pm\!10.6$ & $157.5\!\pm\!11.0$\\
Weight (kg) & $48.9\!\pm\!11.8$ & $51.1\!\pm\!12.2$ & $47.7\!\pm\!10.9$ & $43.7\!\pm\!9.7$\\
BMI (kg/m$^2$) & $18.7\!\pm\!3.1$ & $19.3\!\pm\!3.3$ & $18.2\!\pm\!2.7$ & $17.4\!\pm\!2.6$\\
\bottomrule
\end{tabular*}
\end{table}

Participants with one, two, and three or more walks contributed 569, 606, and 453 recordings, respectively. Individual site cohorts contain 3, 11, 19, 39, 127, 157, 159, 178, and 295 participants, yielding 20 observed combinations of site and recording-count group for the matched ranking audit. Table~\ref{tab:folds} lists per-fold sample sizes, inverse temperatures $t$, and correct participant counts. Within each outer fold, a single regional kernel model (fitted on up to two windows per training participant, selecting $\lambda=0.1$ in every outer fold) is shared across all three random seeds.

\begin{table}[!t]
\caption{Outer-Fold Sizes, Regional Inverse Temperature $t$, and Correct Participant Decisions for the Complete Model.}
\label{tab:folds}\centering\small
\begin{tabular*}{\columnwidth}{@{\extracolsep{\fill}}lrrrrrr@{}}
\toprule
Fold & Fit & Held & $t$ & 73111 & 73112 & 73113\\
\midrule
1 & 789 & 199 & 3.978 & 166 & 170 & 165\\
2 & 789 & 199 & 4.134 & 174 & 173 & 167\\
3 & 791 & 197 & 3.869 & 171 & 170 & 172\\
4 & 791 & 197 & 4.105 & 156 & 154 & 154\\
5 & 792 & 196 & 4.232 & 164 & 163 & 164\\
\midrule
All & & 988 & & 831 & 830 & 822\\
\bottomrule
\end{tabular*}
\end{table}

\section{Participant Errors Across Training Seeds}
Table~\ref{tab:seeds} reports per-seed correct participant counts across model variants. Relative to the frozen-encoder video baseline, the complete system corrects 94 participant--seed errors and introduces 71 (improving two seeds and trailing slightly on the third). Table~\ref{tab:class} breaks these shifts down by screening category; among the 153 missed positive participant--seed instances under the full model, 103 were predicted as negative and 50 as neutral. Table~\ref{tab:cms} provides the full pooled confusion matrix for each seed, with positive recall ranging from 109/162 to 115/162.

\begin{table}[!t]
\caption{Correct Participant Decisions per Training Seed Across the Same 988 Participants.}
\label{tab:seeds}\centering\small
\begin{tabular}{@{}lrrrr@{}}
\toprule
System & 73111 & 73112 & 73113 & Total\\
\midrule
Adapted + regions & 831 & 830 & 822 & 2,483\\
Adapted video & 828 & 828 & 823 & 2,479\\
Frozen + regions & 821 & 823 & 824 & 2,468\\
Frozen video & 819 & 817 & 824 & 2,460\\
\bottomrule
\end{tabular}
\end{table}

\begin{table}[!t]
\caption{Class Recall (\%) and Paired Decision Changes of the Complete Model Relative to the Frozen-Encoder Video Baseline Across Three Seeds.}
\label{tab:class}\centering\small
\begin{tabular}{@{}lrrrr@{}}
\toprule
Category & Full & Frozen & Corrected & Newly Wrong\\
\midrule
Negative & 90.32 & 89.08 & 37 & 17\\
Neutral  & 80.16 & 79.01 & 21 & 11\\
Positive & 68.52 & 69.96 & 36 & 43\\
\bottomrule
\end{tabular}
\end{table}

\begin{table}[!t]
\caption{Confusion Matrices of the Complete 20-Window Model by Seed (Rows: Screening Categories; Columns: Predicted N = Negative, U = Neutral, P = Positive).}
\label{tab:cms}\centering\small
\begin{tabular*}{\columnwidth}{@{\extracolsep{\fill}}llrrrr@{}}
\toprule
Seed & Category & N & U & P & People\\
\midrule
73111 & Negative & 484 & 25 & 28 & 537\\
 & Neutral & 32 & 232 & 25 & 289\\
 & Positive & 30 & 17 & 115 & 162\\
\midrule
73112 & Negative & 488 & 23 & 26 & 537\\
 & Neutral & 33 & 233 & 23 & 289\\
 & Positive & 36 & 17 & 109 & 162\\
\midrule
73113 & Negative & 483 & 24 & 30 & 537\\
 & Neutral & 32 & 230 & 27 & 289\\
 & Positive & 37 & 16 & 109 & 162\\
\bottomrule
\end{tabular*}
\end{table}

\section{Ranking After Matching Acquisition Conditions}
Table~\ref{tab:matched_auc} computes one-versus-rest AUROC from softmax probabilities applied to each participant's 20-window mean logits. A site-by-count stratum contributes to a category's AUROC whenever it contains both in-class and out-of-class participants; pair counts and contributing strata remain identical across seeds.

\begin{table}[!t]
\caption{One-Versus-Rest AUROC Before and After Matching Participants on Site and Recording-Count Group (Mean Across Three Seeds).}
\label{tab:matched_auc}\centering\small
\begin{tabular*}{\columnwidth}{@{\extracolsep{\fill}}lrrrr@{}}
\toprule
Label & All & Matched & Pairs & Groups\\
\midrule
Negative & 0.954 & 0.923 & 5,243 & 7\\
Neutral & 0.955 & 0.894 & 7,518 & 12\\
Positive & 0.943 & 0.819 & 2,365 & 7\\
\midrule
Macro average & 0.950 & 0.879 & & \\
\bottomrule
\end{tabular*}
\end{table}

\section{Sources of Variation in the Patch Gate}
On the 158-participant development split across two seeds (316 observations), clip-conditioned, constant-gate ($g=1/2$), and frozen-encoder video models achieve 264, 263, and 262 correct predictions (mean macro-F1 of 80.46\%, 80.22\%, and 79.73\%, respectively). Across all five held-out folds, 91,054,080 valid patch--frame entries yield a mean gate of $0.439\pm 0.174$, with 61.0\% below $0.5$. Writing $v=e/a$ ($\langle v\rangle=1$ in any nonconstant clip), averaging the algebraic identity
\[
 \frac{v}{1+v}=\frac12+\frac{v-1}{4}
                  -\frac{(v-1)^2}{4(1+v)}
\]
yields the expected-gate formula in the main text, showing how spatially concentrated motion lowers $\langle g\rangle$ below $1/2$ (whereas constant clips default to $1/2$). Because $g_{tr}$ depends solely on observed pixels, every training seed receives identical gate values on any held-out window. After horizontal padding, the 149 real image columns intersect 12 of the 16 patch columns, supplying 192 non-padded or partially padded patches per frame (each normalized by its real-pixel count); across all 19,760 unique scheduled windows, 16.45\% of valid patch--frame gates fall below 0.25 and 2.58\% reach at least 0.75, while downstream impact also depends on the learned difference $U_1-U_0$.

\section{Window Selection and Inference Details}
The fixed 20-slot evaluation bank samples a recording and clip start index with replacement via participant-specific pseudorandom streams, so consecutive slot indices may come from distinct walks. Table~\ref{tab:fixedwindows} reports deterministic subset readouts on zero-based slot 10 ($k=1$), slots $\{5,15\}$ ($k=2$), slots $\{2,6,10,14,18\}$ ($k=5$), odd slots $1,\dots,19$ ($k=10$), and all 20 slots ($k=20$); by contrast, the main-text curve draws 500 independent participant-specific permutations via PCG64 with a fixed random seed shared across all models and training seeds. At inference, regional descriptors are stored in FP16, cast to FP32 and re-normalized prior to kernel evaluation, and accumulated in FP64 across windows; the standalone regional branch classifies 790, 815, 822, 818, and 818 of 988 participants correctly at $k=1,2,5,10,20$ deterministic slots, respectively.

\begin{table}[!t]
\caption{Deterministic Evaluation-Slot Subsets: Accuracy / Macro-F1 (\%).}
\label{tab:fixedwindows}\centering\small
\begin{tabular*}{\columnwidth}{@{\extracolsep{\fill}}lrrr@{}}
\toprule
Windows & Full Model & Adapted Video & Regions Alone\\
\midrule
 1 & 81.07 / 76.67 & 80.53 / 75.91 & 79.96 / 74.61\\
 2 & 84.01 / 80.00 & 83.54 / 79.46 & 82.49 / 78.26\\
 5 & 83.03 / 78.83 & 82.79 / 78.63 & 83.20 / 79.00\\
10 & 84.18 / 80.58 & 83.77 / 80.15 & 82.79 / 78.41\\
20 & 83.77 / 79.91 & 83.64 / 79.82 & 82.79 / 78.67\\
\bottomrule
\end{tabular*}
\end{table}

\section{Class Recall Under Fixed-Weight Input Removal}
Table~\ref{tab:removeclass} details category-specific recall under stream zeroing across all three seeds. Across the 486 positive participant--seed instances, zeroing the silhouette vector while keeping regional scores corrects 49 errors and introduces four, whereas it corrects one error and introduces 26 in the negative class, and corrects three and introduces 20 in the neutral class (yielding 831, 828, and 827 total correct participants at seeds 73111, 73112, and 73113 vs.\ 831, 830, and 822 with both streams active).

\begin{table}[!t]
\caption{Class-Specific Recall (\%) Under Fixed-Weight Stream Removal.}
\label{tab:removeclass}\centering\small\setlength{\tabcolsep}{4pt}
\begin{tabular*}{\columnwidth}{@{\extracolsep{\fill}}llrrr@{}}
\toprule
Video Inputs & Regions & Negative & Neutral & Positive\\
\midrule
RGB + silhouette & Yes & 90.32 & 80.16 & 68.52\\
RGB + silhouette & No & 89.57 & 81.08 & 68.52\\
RGB & Yes & 88.76 & 78.20 & 77.78\\
RGB & No & 88.02 & 78.32 & 77.57\\
Silhouette & Yes & 55.00 & 50.17 & 8.64\\
Silhouette & No & 40.10 & 13.96 & 7.20\\
\bottomrule
\end{tabular*}
\end{table}

\section{Baseline Adaptation Protocols on Scoliosis5K}
GaitSet, GaitPart, and GaitGL trained on silhouettes for 5,940 updates per fold and seed. ScoNet-MT retained its joint identity-triplet and three-class objectives over 20,000 SGD updates (initial learning rate 0.1, momentum 0.9, stepped down at 10,000, 14,000, and 18,000 updates; 30 training frames per sequence and 24 evaluation slots). For FoundationGait-0.13B (whose self-supervised pretraining includes unlabeled Scoliosis1K walks already excluded during Scoliosis5K construction), $64\times64$ silhouettes were center-cropped to $64\times44$: linear probing pooled eight training windows per participant into 15,872-dimensional descriptors (AdamW at $10^{-3}$ for 60 epochs), whereas supervised fine-tuning sampled four 30-frame clips from each of eight participants per batch and updated the last two backbone stages, projector, and head with SGD (learning rates 0.001, 0.005, 0.01, and 0.1, momentum 0.9, weight decay $5\times10^{-4}$, label smoothing, $10\times$ decay at 2,000 updates, evaluated at 3,000 updates). FoundationGait-0.03B trained a linear probe for 3,000 updates (sampling 30 frames from 60-frame intervals across eight clips for each of eight participants per batch) followed by 2,000 fine-tuning updates (34-frame intervals; learning rates 0.001 on the last two stages, 0.01 on projector/predictor, and 0.1 on the classifier, with batch-norm running statistics updated throughout).

For cropped RGB optical flow, RAFT-Large estimated displacements between consecutive stored crops resized to $128\times256$ ($W\times H$), normalized horizontally and vertically by width and height, and resized to $44\times64$ for ScoNet-MT (30 training flow pairs per sequence). ScoliGaitX fused $64\times44$ silhouette, derived optical-flow, and gait-energy-residual sequences via gated fusion (triplet margin 1.5, GeM exponent 6.5), while TG-MILNet formed four DTW temporal bags per window paired with five fixed CLIP text prompts (both trained with Adam at $10^{-4}$ for 100 epochs). For directional cross-modal attention (MulT), models trained for 40 epochs per fold with inner-fold learning-rate selection over the paired RGB and silhouette streams.

\section{Paired Teacher Controls}
Swapping the RGB teacher target across participants sharing the same screening category and window slot yields paired-minus-donor macro-F1 differences of $+0.76$, $-0.06$, and $+0.73$ percentage points across the three seeds (mean $+0.48$ percentage points), whereas regressing leave-one-out class-mean targets reaches 79.15\% macro-F1 compared with 78.85\% (82.69\% accuracy) without a teacher.

\section{Multi-Window Readouts and Regional Inference Cost}
Table~\ref{tab:windows} lists mean accuracy and macro-F1 across the 500 random-permutation draws for $k\in\{1,2,5,10,20\}$ windows. Each evaluation window executes up to 24 DINOv2 frame passes plus one PE-Spatial pass, and regional kernel evaluation computes $16n$ cell comparisons across $n$ training participants (1,578 to 1,584 reference windows per fold, shared across seeds; the 73,728 adapter parameters count only the low-rank RGB updates alongside both frozen vision backbones and the trained silhouette transformer and fusion head).

\begin{table}[!t]
\caption{Mean Accuracy / Macro-F1 (\%) Across 500 Random Evaluation-Slot Draws.}
\label{tab:windows}\centering\small
\begin{tabular*}{\columnwidth}{@{\extracolsep{\fill}}lrrr@{}}
\toprule
Windows & Full Model & Adapted Video & Regions Alone\\
\midrule
 1 & 82.05 / 77.77 & 81.60 / 77.21 & 79.77 / 74.83\\
 2 & 82.89 / 78.79 & 82.47 / 78.28 & 81.39 / 76.77\\
 5 & 83.49 / 79.58 & 83.12 / 79.13 & 82.41 / 78.16\\
10 & 83.70 / 79.85 & 83.39 / 79.47 & 82.73 / 78.62\\
20 & 83.77 / 79.91 & 83.64 / 79.82 & 82.79 / 78.67\\
\bottomrule
\end{tabular*}
\end{table}
\end{document}